# A Comparative Study on Basic Elements of Deep Learning Models for Spatial-Temporal Traffic Forecasting

Yuyol Shin[1][0000-0002-1445-0112], and Yoonjin Yoon[1][0000-0002-3550-4431](✉)

[1] Korea Advanced Institute of Science and Technology, 34141, Daejeon, South Korea
`yoonjin@kaist.ac.kr`

**Abstract.** Traffic forecasting plays a crucial role in intelligent transportation systems. The spatial-temporal complexities in transportation networks make the problem especially challenging. The recently suggested deep learning models share basic elements such as graph convolution, graph attention, recurrent units, and/or attention mechanism. In this study, we designed an in-depth comparative study for four deep neural network models utilizing different basic elements. For base models, one RNN-based model and one attention-based model were chosen from previous literature. Then, the spatial feature extraction layers in the models were substituted with graph convolution and graph attention. To analyze the performance of each element in various environments, we conducted experiments on four real-world datasets - highway speed, highway flow, urban speed from a homogeneous road link network, and urban speed from a heterogeneous road link network. The results demonstrate that the RNN-based model and the attention-based model show a similar level of performance for short-term prediction, and the attention-based model outperforms the RNN in longer-term predictions. The choice of graph convolution and graph attention makes a larger difference in the RNN-based models. Also, our modified version of GMAN shows comparable performance with the original with less memory consumption.

**Keywords:** Traffic Forecasting, Graph Convolutional Networks, Graph Attention Networks, Intelligent Transportation System, Urban Traffic Dataset.

## 1    Introduction

Traffic forecasting has been an actively studied problem in the transportation research area for the past four decades. An accurate forecast of traffic can help develop various applications such as network capacity evaluation, travel time estimation, and signal optimization. It is a crucial technical enabler of Intelligent Transportation Systems (ITS), but capturing the spatial-temporal complexities in transportation road networks is not a simple task.

Traditionally, data-driven approaches to traffic forecasting have included simple time series models such as ARIMA (Auto-Regressive Integrated Moving Average) [1], VAR (Vector Auto-Regressive model) [2], and Kalman filtering [3]. Although these traditional models had advantages in interpretability of the model parameters, such models had limited ability on modeling the spatial-temporal complexities. The majority of studies focused on forecasting traffic measurements based on limited sets of sensors in expressways.

The deep learning models have demonstrated notable improvement in capturing the spatial-temporal complexities, especially in forecasting accuracy and scale. Some of the early works employed Recurrent Neural Networks (RNN) [4-6] and Convolutional Neural Networks (CNN) [7-8]. More recently, Graph Neural Networks (GNN) [9] has gained popularity in the field to incorporate the graph structures inherent in transportation networks. The most prevalent form of the traffic forecasting model is the convolutional GNN layers combined with RNNs [10-13], which showed promising results compared to the deep learning models that did not consider the spatial connection of road network graphs. Some efforts have been made to model the temporal complexity with gated convolutions [14-17]. These models showed advantages in computation time since sequential computation required for RNNs was not necessary. Recently, attention [18] has gained popularity in both the temporal [19] and spatial [20] domains. Cai et al. [19] showed utilizing transformer [18] architecture can be helpful. Zhang et al. [20] introduced gated attention to aggregate neighbor information in graphs. Combined with Gated Recurrent Units (GRU) [21], the attentional GNN layer showed improved performance over convolutional GNN layers. Instead of aggregating information from neighboring nodes, Zheng et al. [22] have adopted embedding techniques of nodes and graphs.

The deep learning models have indeed thrived with their ability to produce high accuracy results in large-scale network datasets. However, there have not been sufficient efforts to understand the models in detail. While many studies utilize GNN layers such as the Graph Convolutional Networks (GCN) [23], Graph Attention Networks (GAT) [24], and Diffusion Convolution [10] to extract spatial features, there is a significant lack of comparative studies of such models. Li et al. [10] proposed a graph convolution operation for traffic forecasting, named Diffusion Convolution, by relating traffic flow to a diffusion process. In the study, they made a comparison between the Diffusion Convolution and ChebNet [25] for overall performance on one of their study areas. Cui et al. [11] made a comparison among the proposed Traffic Graph Convolution (TGC), spectralCNN [26], and ChebNet [25], in terms of the number of parameters, computation time, and ability to extract localized features along with the overall performance and rate of convergence. Even though the aforementioned studies provided comparisons between the new layers and the existing ones, an in-depth comparative study on various models is necessary to justify and understand the performances of the traffic forecasting models beyond incremental change. Furthermore, an evaluation of layers for the temporal feature extraction (RNNs and attention) is also necessary.

In this study, we conducted extensive experiments for an in-depth results analysis to characterize each basic element of traffic forecasting models. Specifically, we chose two models – T-GCN [12] & GMAN [22] – based on how they extract the temporal features. T-GCN is an RNN-based model that combines GCN and GRU [21], and GMAN is an attention-based model that implements gated fusion to combine temporal attention and spatial attention. Then, we tested the models by trying both GCN and GAT for spatial feature extraction on four different datasets. The results show that the attention-based model outperforms the RNN-based model in longer-term prediction, and produces a similar amount of error for short-term prediction. Moreover, the choice of spatial feature extraction layer affects the performance more in the attention-based model. The main contributions of our work are summarized as follows.

- We conducted extensive experiments and in-depth analysis on the basic elements of traffic forecasting models considering both spatial and temporal characteristics – GCN and GAT for spatial feature extraction, and RNN and attention for temporal feature extraction. We discovered the trends of the forecasting results of each basic element.
- When we substituted the spatial attention layer in GMAN to GAT and GCN, the results show that the GAT-substituted model produces a similar level of performance with less memory consumption.
- We published two speed datasets of road segments in Seoul, South Korea.[1] It is a rare dataset which contains speed generated in highly urbanized environments where various characteristics such as nodes with high degrees, signalization, and different land uses make the traffic forecasting problem even more complex.

## 2   Method & Data

In this section, we first make the mathematical definition of the traffic forecasting problem. The GNN-based spatial feature extraction methods are explained along with the base models. Finally, four real-world datasets used in this study are introduced. The outline of this comparative study is in figure 1.

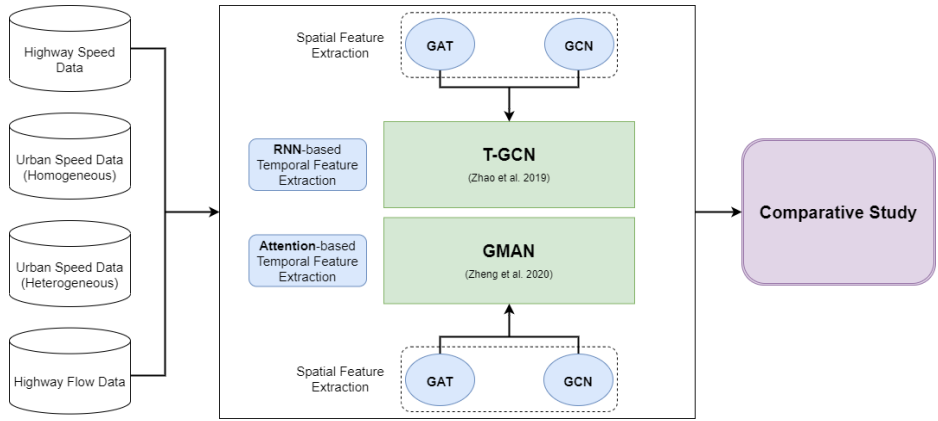

**Fig 1.** Overall framework of this comparative study

---

[1] https://github.com/yuyolshin/SeoulSpeedData

## 2.1 Problem Definition

A graph for transportation network is defined as $\mathcal{G} = (V, E)$ where $V$ is the set of nodes with $|V| = N$ nodes and $E$ is the set of edges representing the connectivity of the nodes. The adjacency matrix $\mathbf{A} \in \mathbb{R}^{N \times N}$ is derived from the graph where each element $a_{ij} \geq 0$ if $(v_i, v_j) \in E$.

Graph signal at time $t$ is $\mathbf{X}_t = \{x_t^1, \dots, x_t^N\} \in \mathbb{R}^{N \times C}$ where $C$ is the number of features for the signal. Given a graph at time $t$, the traffic forecasting problem is to predict the future traffic state for $T_p$ time steps using signals of $T_h$ historical time steps.

$$H: [X_{t-T_h+1}, \dots, X_t] \to [\widehat{Y}_{t+1}, \dots, \widehat{Y}_{t+T_p}] \tag{1}$$

## 2.2 Graph Neural Networks

Most GNN layers can be categorized into three groups: convolutional, attentional, and message-passing. These techniques differ in how each node aggregates information from its neighboring nodes. In graph convolution, source node features are multiplied by a scalar weight. A graph attention layer calculates the scalar weight with a function of target and source node features. Finally, vector-based messages are calculated with a function of target and source node features in message-passing. For traffic forecasting, convolutional and attentional GNNs are more popular than message-passing GNNs.

In this study, we conducted experiments with one convolutional and one attentional GNN layers – namely, GCN [23] and GAT [24]. The GCN layer constructs filter in the Fourier domain using first order approximation of Chebyshev polynomials, having the ability to extract spatial features by aggregating information from the nodes of a graph and their first-order neighbors. Moreover, the relationship between the nodes and neighbors of higher orders can be captured by stacking multiple GCN layers. A GCN layer applied on the input $\mathbf{X} \in \mathbb{R}^{N \times d}$ of $N$ nodes of $d$ dimensions can be expressed as:

$$\text{GCN}(\mathbf{X}, \mathbf{A}) = \sigma(\widehat{\mathbf{A}} \mathbf{X} \mathbf{W}), \tag{2}$$

where $\sigma(\cdot)$ is an activation function, $\widehat{\mathbf{A}} = \widetilde{\mathbf{D}}^{-1/2} \widetilde{\mathbf{A}} \widetilde{\mathbf{D}}^{-1/2}$ is the normalized Laplacian matrix where $\widetilde{\mathbf{A}} = \mathbf{I}_N + \mathbf{A}$, and $\widetilde{\mathbf{D}}_{ii} = \sum_j \tilde{a}_{ij}$, and $\mathbf{W} \in \mathbb{R}^{d \times h}$ is the weight parameter matrix where $h$ is the dimension of the output layer.

The GAT layer employs the multi-head self-attention mechanism suggested by [16] in node classification tasks. In the layer, the queries are the source and target node features, keys are learnable parameters, and values are the source node features. The attention score between node $i$ and node $j$ for head $k$ can be obtained as:

$$\alpha_{ij}^k = \frac{\exp\left(\sigma\left(\mathbf{a}^{k^T} \text{CAT}(\mathbf{W}^k \mathbf{x}_i, \mathbf{W}^k \mathbf{x}_j)\right)\right)}{\sum_{v \in \mathcal{N}_i} \exp\left(\sigma\left(\mathbf{a}^{k^T} \text{CAT}(\mathbf{W}^k \mathbf{x}_i, \mathbf{W}^k \mathbf{x}_v)\right)\right)}, \tag{3}$$

where $\mathbf{a}^k \in \mathbb{R}^{2h'}$ is a single layer feedforward neural network for attention head $k$, where $h'$ is the dimension of each head, $\mathbf{W}^k \in \mathbb{R}^{d \times h'}$ is the learnable weight parameter matrix for attention head $k$, $\mathcal{N}_i$ is the neighbor set of node $i$, and $\text{CAT}(\cdot)$ is the concatenation operation.

The GAT layer with $K$ heads applied on the input $\mathbf{x}_i \in \mathbb{R}^{N \times d}$ can be expressed as:

$$\text{GAT}(x_i, \mathbf{A}) = \text{CAT}_{k=1}^{K}\left(\sigma\left(\sum_{v \in \mathcal{N}_i} \alpha_{ij}^k \mathbf{W}^k x_v\right)\right). \tag{4}$$

When we set the output dimension for GAT $h' \times k$ to $h$, the output dimensions for both the GCN and GAT are in $\mathbb{R}^{N \times h}$. Thus, switching one from the other is possible for any given traffic forecasting model.

### 2.3 Base Models

The purpose of this study is to compare the traffic forecasting results for RNN/attention-based models and GCN/GAT-based models and discover the characteristics of each basic element. We choose one RNN-based model (T-GCN [12]), and one attention-based model (GMAN [22]) from previous literature as our base model and conduct experiments by changing their spatial feature extraction layer to GCN and GAT.

T-GCN is a traffic forecasting model based on GRU and GCN. The model uses a 2-layer GCN model combined with GRU units. The GRU units for the model are defined as:

$$\boldsymbol{u_t} = \sigma(\mathbf{W}_u[f(\mathbf{A}, \mathbf{X}_t), \boldsymbol{h}_{t-1}] + \boldsymbol{b}_u) \tag{5}$$
$$\boldsymbol{r_t} = \sigma(\mathbf{W}_r[f(\mathbf{A}, \mathbf{X}_t), \boldsymbol{h}_{t-1}] + \boldsymbol{b}_r) \tag{6}$$
$$\boldsymbol{c}_t = \tanh(\mathbf{W}_c[f(\mathbf{A}, \mathbf{X}_t), (\boldsymbol{r}_t * \boldsymbol{h}_{t-1})] + \boldsymbol{b}_c) \tag{7}$$
$$\boldsymbol{h}_t = \boldsymbol{u}_t * \boldsymbol{h}_{t-1} + (1 - \boldsymbol{u}_t) * \boldsymbol{c}_t \tag{8}$$

where $\sigma$ is the sigmoid activation function, $f(\mathbf{A}, \mathbf{X}_t) = \sigma(\hat{\mathbf{A}} \text{ReLU}(\hat{\mathbf{A}} \mathbf{X} \mathbf{W}_0) \mathbf{W}_1)$ is the 2-layer GCN model where $\mathbf{W}_0 \in \mathbb{R}^{d \times p}$, and $\mathbf{W}_1 \in \mathbb{R}^{p \times h}$ are learnable parameter matrices, and $\mathbf{W}_u, \mathbf{W}_r, \mathbf{W}_c \in \mathbb{R}^{2h \times h}$, and $\boldsymbol{b}_u, \boldsymbol{b}_r, \boldsymbol{b}_c \in \mathbb{R}^h$ are the learnable parameter matrices and biases, respectively. We used T-GCN and T-GAT which replaces the GCN layer to GAT as experiment models.

GMAN uses multi-head self-attention in both temporal and spatial domains. The model separately extracts spatial and temporal features and combines them through gated fusion. Instead of the positional encoding used in Vaswani et al. [18], they suggest the spatial-temporal embedding using time indicator vectors and node embedding vectors obtained by node2vec [27]. In GMAN, the spatial feature vector $hs_{i,t}^{(l)}$ for node $i$ on time $t$ is defined as:

$$hs_{i,t}^{(l)} = \text{CAT}_{k=1}^{K}\left(\sum_{v \in V} \alpha_{i,v}^{(k)} \cdot f_{s,0}^{(k)}(h_{v,t}^{(l-1)})\right) \tag{9}$$

where $\alpha_{i,v}^{(k)}$ is the attention score for head $k$, $f_{s,0}^{(k)}$ is a non-linear projection of previous hidden state, $hs^{(l)}$ is the state vector after $l$-th spatial attention layer, and $h^{(l-1)}$ is the state vector after $(l-1)$-th GMAN layer. The attention score $\alpha_{i,v}^{(k)}$ between node $i$ and $v$ for head $k$ can be obtained through:

$$s_{i,v}^{(k)} = \frac{\langle f_{s,1}^{(k)}\left(\text{CAT}(h_{i,t}^{(l-1)}, e_{i,t})\right), f_{s,2}^{(k)}\left(\text{CAT}(h_{v,t}^{(l-1)}, e_{v,t})\right)\rangle}{\sqrt{d}} \tag{10}$$

$$\alpha_{i,j}^{(k)} = \frac{\exp(s_{i,j}^{(k)})}{\sum_{v \in V} \exp(s_{i,v}^{(k)})} \tag{11}$$

where $f_{s,1}^{(k)}$ and $f_{s,2}^{(k)}$ are non-linear projections, $e_{i,t}$ is the embedding vector at time $t$ on node $i$, $\langle \cdot, \cdot \rangle$ is inner product operator, and $d$ is the dimension of output. In our experiments, we replace the spatial attention layer for the model with GCN (GMAN-GCN) and GAT (GMAN-GAT) layers, so that the layer only aggregates information from neighboring nodes instead of calculating attentions of all possible pairs. The temporal attention is calculated in a similar way to the spatial attention. The transform attention is achieved through calculating attention scores between the historical time steps and the prediction time steps. Lastly, the gated fusion is implemented with two non-linear projections. For detailed model descriptions, we refer the readers to Zheng et al. [22].

## 2.4 Data

To analyze the performance of each basic element, we select four real-world datasets that possess different characteristics of traffic data: the PeMS-Bay, METR-LA [10], Urban-core, and Urban-mix datasets [13].

The PeMS-Bay dataset is a widely used speed dataset for traffic forecasting collected by California Transportation Agencies (CalTrans) Performance Measurement System (PeMS). The dataset contains 6 months of data ranging from January 1st, 2017 to June 30th, 2017 with a data frequency of 5 minutes. Spatially, 325 sensors in the Bay Area are included. We selected this dataset to examine the performance of the models for loop detector-based highway speed forecasting.

The METR-LA dataset is a traffic flow dataset containing the data collected from loop detectors in the highway of Los Angeles County. It is another dataset that has been frequently used in traffic forecasting studies. The dataset contains 5-minute traffic flow data from 207 sensors and ranges from Mar 1st, 2012 to June 30th, 2012 (4 months). Through the experiments on this dataset, we analyzed the differences between the model performances on traffic speed and flow datasets

The Urban-core and Urban-mix datasets are processed from the Seoul traffic speed dataset distributed by Transport Operation & Information Service (TOPIS). The dataset aggregates the signals from DTG (Digital Tacho Graph) on Seoul taxis into 5-minute speed data for road segments in the Seoul traffic network. The original dataset contains 4,774 road segments for one month ranging from April 1st, 2018 to April 30th, 2018. The Urban-core dataset contains 304 road segments in Gangnam, Seoul, South Korea, one of the areas with the highest traffic in the country. The road segments in the site have identical speed limits and similar degrees, and segment lengths. The Urban-mix dataset is an expansion of the Urban-core dataset and includes 1,007 road segments with more diverse characteristics. Compared to the highway datasets, both of the Seoul speed datasets show more complexity as in figure 2. Approximate entropy [28] of the sensors (and road segments) on average are 0.52, 1.20, 1.40, and 1.41 for PeMS-Bay, METR-LA, Urban-core, and Urban-mix, respectively. We used these two datasets to analyze the model performances in complex urban environments.

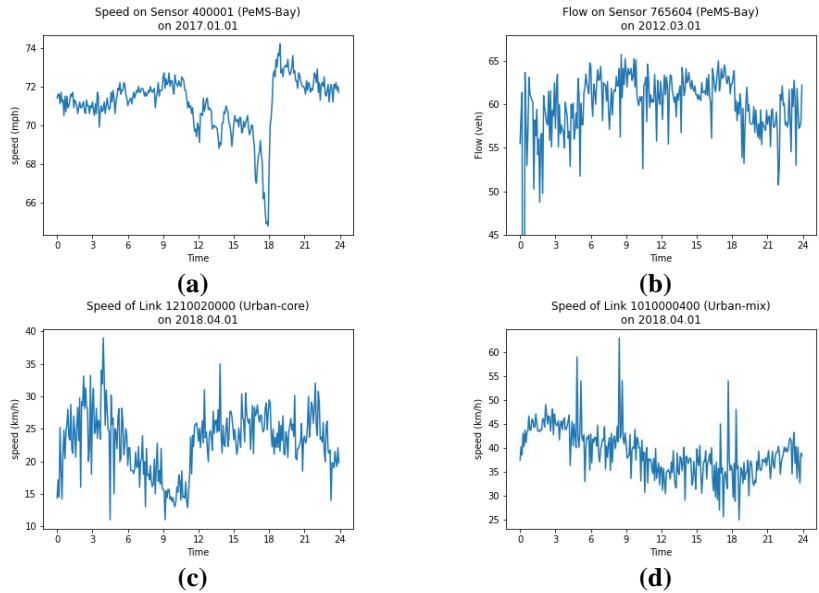

**Fig 2.** One day sample data from different datasets. **(a)** PeMS-Bay, **(b)** METR-LA, **(c)** Urban-core, and **(d)** Urban-mix

**Data Processing.** For the PeMS-Bay and METR-LA datasets, we followed the procedures used in Li et al. [10]. The train, validation, test sets are divided by the proportion of 70%, 10%, and 20%, respectively. Each element $a_{ij}$ of the adjacency matrix A for the traffic sensors are defined as:

$$a_{ij} = f(x) = \begin{cases} \exp\left(-\frac{d_{ij}^2}{\sigma^2}\right), & \text{if } \exp\left(-\frac{d_{ij}^2}{\sigma^2}\right) \geq \epsilon \\ 0, & \text{otherwise} \end{cases}, \quad (10)$$

where $d_{ij}$ is the distance between sensor $i$ and sensor $j$, $\sigma$ is the standard deviation, and $\epsilon$ is the threshold value, which is set to 0.1.

For the Urban-core and Uran-mix datasets, the training, validation, and test sets are divided into 21, 2, and 7 days, respectively. Table 1 summarizes the datasets used in this study.

**Table 1.** Summary of the dataset

|  | PeMS-Bay | METR-LA | Urban-core | Urban-mix |
|---|---|---|---|---|
| # Nodes | 325 | 207 | 304 | 1,007 |
| # Edges | 2,694 | 1,722 | 1,696 | 5,597 |
| Resolution | 5-min | 5-min | 5-min | 5-min |
| Duration | 6 months | 4 months | 1 month | 1 month |
| Site | Express | Express | Urban | Urban |
| Sensor Type | Loop | Loop | GPS | GPS |

## 3 Experiment

### 3.1 Metrics

For the evaluation of the models in different areas, we adopt mean absolute error (MAE), root mean squared error (RMSE), and mean absolute percentage error (MAPE). Since the flow dataset contains 0 values as its true value, MAPE is not calculated for the METR-LA dataset. The evaluation metrics are defined as the following:

$$\text{MAE} = \frac{1}{T_p N} \sum_{i=1}^{N} \sum_{j=1}^{T_p} |\widehat{Y}_{ij} - Y_{ij}|, \qquad (11)$$

$$\text{RMSE} = \sqrt{\sum_{i=1}^{N} \sum_{j=1}^{T_p} \frac{(\widehat{Y}_{ij} - Y_{ij})^2}{T_p N}}, \qquad (12)$$

$$\text{MAPE} = \frac{1}{T_p N} \sum_{i=1}^{N} \sum_{j=1}^{T_p} \frac{|\widehat{Y}_{ij} - Y_{ij}|}{Y_{ij}}. \qquad (13)$$

where $T_p$ is the total number of predicted time steps, $N$ is the number of nodes (sensors or road segments), $\widehat{Y}_{ij}$ is the predicted value, and $Y_{ij}$ is the actual value.

### 3.2 Hyperparameters

Hyperparameters are set as closely as indicated in the original works [12, 22]. For all models, we set the hidden units to 64, batch size to 32, and learning rate to 0.001. For GAT, both the number of heads and the dimensions of each head are 8. The number of layers for GMAN models is 3 except for the Urban-mix dataset because of memory consumption. For the Urban-mix dataset, we use a 2-layer model. We train the models using Adam optimizer and the early stopping with a patience of 10.

### 3.3 Overall Performance Comparison

In this section, we compare the overall performance of each model in the four datasets. Table 2 shows the results of experiments on the four datasets for 15 minutes (3 steps), 30 minutes (6 steps), and 1 hour (12 steps). For the RNN-based model, T-GAT yields more accurate performances than T-GCN except for MAE on 15-minute prediction of the METR-LA dataset. For the attention-based model, GMAN-GAT generally produces better performance in the speed datasets, and GMAN-GCN yields more accurate results in the flow dataset. The choice of GNN layers affects the performance more for the RNN-based model. By substituting GCN to GAT for 1-hour prediction in the PeMS-Bay dataset, RMSE changes by 3.81% for the attention-based model, and 1.31% for the RNN-based model. Also, the RNN-based model shows competitiveness for short-term traffic forecasting and traffic flow forecasting. T-GAT shows the best performance for 15-minute prediction in the Urban-core dataset and RMSE values in the METR-LA dataset for all prediction horizons.

9**Table 2.** Overall performance comparison for the models

| Prediction | Metric | T-GCN | T-GAT | GMAN-GCN | GMAN-GAT |
|---|---|---|---|---|---|
| **PeMS-Bay** | | | | | |
| 15 min | MAE | 2.25 | 1.58 | 1.49 | **1.43** |
| | RMSE | 3.78 | 2.96 | 2.99 | **2.89** |
| | MAPE | 4.74 | 3.40 | 3.41 | **3.08** |
| 30 min | MAE | 2.51 | 1.98 | 1.78 | **1.73** |
| | RMSE | 4.46 | 3.95 | 3.78 | **3.73** |
| | MAPE | 5.54 | 4.47 | 4.20 | **3.91** |
| 60 min | MAE | 2.91 | 2.46 | 2.06 | **2.02** |
| | RMSE | 5.25 | 5.05 | 4.42 | **4.37** |
| | MAPE | 6.69 | 5.90 | 5.00 | **4.70** |
| **METR-LA** | | | | | |
| 15 min | MAE | **4.53** | 4.55 | 5.03 | 5.32 |
| | RMSE | 9.48 | **9.26** | 11.50 | 11.69 |
| 30 min | MAE | 6.08 | 5.91 | **5.88** | 6.12 |
| | RMSE | 11.88 | **11.56** | 13.20 | 13.30 |
| 60 min | MAE | 8.09 | 7.56 | **7.16** | 7.36 |
| | RMSE | 14.56 | **14.28** | 15.29 | 15.33 |
| **Urban-core** | | | | | |
| 15 min | MAE | 2.91 | **2.62** | 2.69 | **2.62** |
| | RMSE | 4.12 | **3.87** | 3.99 | 3.94 |
| | MAPE | 11.85 | **10.21** | 10.93 | 10.69 |
| 30 min | MAE | 3.10 | 2.92 | 2.73 | **2.69** |
| | RMSE | 4.38 | 4.20 | 4.05 | **4.00** |
| | MAPE | 12.85 | 11.55 | 11.14 | **10.90** |
| 60 min | MAE | 3.46 | 3.26 | 2.86 | **2.82** |
| | RMSE | 4.81 | 4.60 | 4.18 | **4.13** |
| | MAPE | 14.57 | 13.18 | 11.71 | **11.48** |
| **Urban-mix** | | | | | |
| 15 min | MAE | 3.37 | 2.92 | 2.93 | **2.86** |
| | RMSE | 4.90 | **4.45** | 4.53 | 4.46 |
| | MAPE | 13.29 | **10.84** | 11.22 | 11.09 |
| 30 min | MAE | 3.63 | 3.47 | 3.10 | **3.05** |
| | RMSE | 5.43 | 5.23 | 4.86 | **4.83** |
| | MAPE | 14.42 | 12.81 | 11.94 | **11.91** |
| 60 min | MAE | 4.90 | 4.45 | 4.53 | **4.46** |
| | RMSE | 6.12 | 6.09 | 5.40 | **5.39** |
| | MAPE | 16.42 | 15.00 | **13.15** | 13.19 |

### 3.4 Robustness against noises

The real-world traffic data is prone to noises. In our experiments, we observed different levels of robustness against noises between the RNN-based model and the attention-based model. For the 1-hour forecasts of the PeMS-Bay dataset, both T-GCN and T-GAT show sudden drops when there are noisy data in the label. The T-GCN forecast shows a delayed start of peak hours, and the T-GAT forecast shows a delayed end of peak hours (Figure 3(a) and (b)). In the Urban-core dataset, where noises are larger than PeMS-Bay, the forecasts of the RNN-based models also show delayed trends of the actual data (Figure 3(c) and (d)).

For the flow dataset, T-GAT still makes sudden drops in forecast values due to noises. However, the overall trend is better captured by T-GAT than by GMAN-GAT (Figure 4).

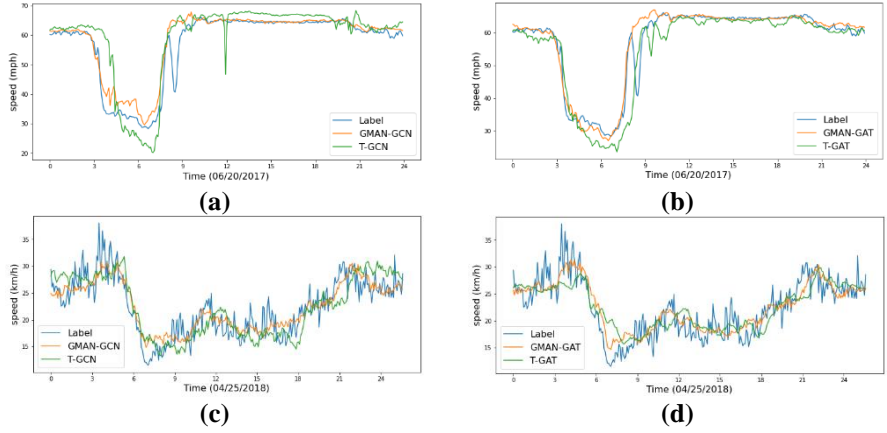

**Fig. 3. (a)** and **(b)** are 1-hour prediction on sensor 401327 in the Bay area (PeMS-Bay). **(c)** and **(d)** are 1-hour prediction on Link 1220019000 in Urban-core.

### 3.5 Comparison between GMAN and GMAN-GAT

Table 3 presents the performance of GMAN and GMAN-GAT on the PeMS-Bay dataset. Although GMAN produces more accurate forecasts, GMAN-GAT offers an advantage in memory consumption compared to GMAN. Since GMAN calculates the attention score between all node pairs, the required number of attention scores to calculate is $N^2$. With grouped spatial attention, the number can be reduced to $2^{-1/3}N^{4/3}$. By using GAT instead of spatial attention, the number of attention scores is the number of edges $|E|$, which is generally smaller than $2^{-1/3}N^{4/3}$ in most graphs including the datasets used in this study.

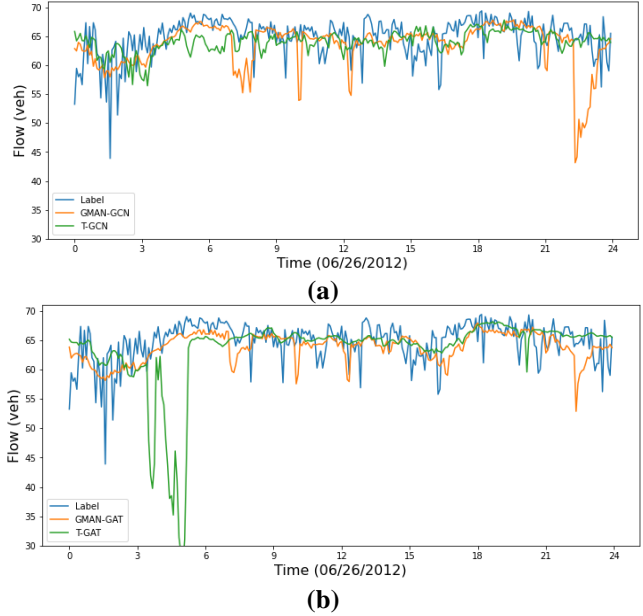

**Fig. 4.** 1-hour prediction values in the METR-LA dataset, **(a)** using graph convolution layers, and **(b)** graph attention layers

**Table 3.** The performance of GMAN and GMAN-GAT on PeMS-Bay on 1-hour prediction

|  | MAE | RMSE | MAPE |
| --- | --- | --- | --- |
| GMAN | 1.86 | 4.32 | 4.31 |
| GMAN-GAT | 2.02 | 4.37 | 4.70 |

## 4    Conclusion and Future Study

In this study, we conducted an in-depth comparative study for the basic elements of deep learning models for traffic forecasting using four different real-world datasets. GCN and GAT layers for the spatial feature extractions and RNN and attention for the temporal feature extractions were analyzed. The results show that no model outperforms the rest in all scenarios. The attention-based models outperform the RNN-based models in longer-term speed prediction. For traffic flow forecasting, however, the RNN-based model shows competitive performance in all prediction horizons. GMAN-GAT shows a more accurate forecast in the speed prediction, but for the flow prediction, GMAN-GCN outperforms GMAN-GAT. The attention-based model shows robustness against noises, but T-GAT fits better to the overall trend.

In future work, we will conduct experiments with more combinations of basic elements such as diffusion convolution [10], multi-weight traffic graph convolution [13], and adaptive graph convolution [16], and define the characteristics of each element. To uncover the characteristics of each basic elements, we plan to adopt explainable AI techniques [29-30], which is rarely used in the field of traffic forecasting research [31]. Finally, the state-of-the-art models [16-17, 22] will be analyzed to find out whether the characteristics of the basic elements would remain in more sophisticated architectures. The purpose of future study is to uncover the black-box nature of the deep learning models for transportation and open the opportunities for the models to be used in real-world applications.

## 5  Acknowledgements

This work was supported by the National Research Foundation of Korea (NRF) Basic Research Lab grant (2020R1A2C2010200) and Midcareer Research Grant (2021R1A4A1033486) by South Korean government.